\documentclass[11pt,a4paper]{article}
\usepackage[hyperref]{emnlp-ijcnlp-2019}
\usepackage{times}
\usepackage{latexsym}
\usepackage{graphicx}
\usepackage{amssymb,amsmath,amsthm}
\usepackage{xcolor}
\usepackage{array}
\usepackage{setspace}

\usepackage{url}
\usepackage{comment}

\usepackage{pifont}

\newtheoremstyle{mytask}%
{}{}{\normalfont}{}{\bf}{ }{ }{}
\theoremstyle{mytask}
\newtheorem{task}{Task}

\aclfinalcopy 

\title{Abstractive Summarization with Combination of Pre-trained Sequence-to-Sequence
and Saliency Models\thanks{\ \ Work in progress.}}

\author{Itsumi Saito, Kyosuke Nishida, Kosuke Nishida, Junji Tomita \\
  NTT Media Intelligence Laboratory, NTT Corporation \\
  {\tt itsumi.saito.df@hco.ntt.co.jp} \\
  }

\date{}

\begin{document}
\maketitle
\begin{abstract}
Pre-trained sequence-to-sequence (seq-to-seq) models have significantly improved the accuracy of several language generation tasks, including abstractive summarization. 
Although the fluency of abstractive summarization has been greatly improved by fine-tuning these models, it is not clear whether they can also identify the important parts of the source text to be included in the summary. 
In this study, we investigated 
the effectiveness of combining saliency models that identify the important parts of the source text with the pre-trained seq-to-seq models through extensive experiments.
We also proposed a new combination model consisting of a saliency model that extracts a token sequence from a source text and a seq-to-seq model that takes the sequence as an additional input text.
Experimental results showed that 
most of the combination models outperformed a simple fine-tuned seq-to-seq model on both the CNN/DM and XSum datasets  
even if the seq-to-seq 
model is pre-trained on large-scale corpora. 
Moreover, for the CNN/DM dataset, the proposed combination model 
exceeded the previous best-performed model by 1.33 points on ROUGE-L.
\end{abstract}

\section{Introduction}
Pre-trained language models such as BERT~\cite{bert} have significantly improved the accuracy of various language processing tasks. However, we cannot apply BERT to language generation tasks as is because its model structure is not suitable for language generation. Several pre-trained seq-to-seq models for language generation~\cite{bart,2019t5}  based on an encoder-decoder Transformer model, which is a standard model for language generation, have recently been proposed. These models have achieved \textcolor{black}{state-of-the-art} results in various language generation tasks, including abstractive summarization. 

However, when generating a summary, it is essential to correctly predict which part of the source text should be included in the summary. Some previous studies without pre-training have examined combining extractive summarization with abstractive summarization~\cite{bottomup, unified}. Although pre-trained seq-to-seq models 
have achieved higher accuracy compared to previous models, it is not clear whether modeling ``Which part of the source text is important?'' can be learned through pre-training.  

\textcolor{black}{The purpose of this study is to clarify the }
\textcolor{black}{effectiveness of combining saliency models that identify the important part of the source text with a pre-trained seq-to-seq model in the abstractive summarization task. }
Our main contributions are as follows:

\begin{itemize}
  \item 
  We investigated 
  nine combinations of 
  pre-trained seq-to-seq and token-level saliency models, where the saliency models \textit{share} the parameters with the encoder of the seq-to-seq model or \textit{extract} important tokens independently of the encoder.
 \item 
 We proposed a new combination model, the conditional summarization model with important tokens (\textbf{CIT}), in which a token sequence extracted by a saliency model is explicitly given to a seq-to-seq model as an \textit{additional} input text. 
 \item
 We evaluated 
 the combination models
 on the CNN/DM~\cite{cnndm} and XSum~\cite{xsum} datasets. Our CIT model outperformed a simple fine-tuned model 
 in terms of ROUGE scores on both datasets. 
\end{itemize}

\section{Task Definition}
Our study focuses on two tasks: abstractive summarization and 
\textcolor{black}{saliency detection.} 
The main task is abstractive summarization and the sub task is 
\textcolor{black}{saliency detection,} 
which is the prediction of important parts of the source text. The problem formulations of each task are described below.

\begin{task}
[Abstractive summarization] Given the source text $X$, 
the output is 
an abstractive summary $Y$ = $(y_1,\ldots,y_T)$. 
\end{task}

\begin{task}
[Saliency detection] Given the source text $X$ with $L$ words $X$= $(x_1,\dots,x_L)$, 
the output is 
the saliency score $S = \{S_1, S_2, ... S_L \}$. 
\end{task}

In this study, we investigate several combinations of models for these two tasks.

\section{Pre-trained seq-to-seq Model}
\label{sec:seq-to-seq}

There are several pre-trained seq-to-seq models applied for abstractive summarization~\cite{mass, unilm, 2019t5}. 
The models use
a simple Transformer-based encoder-decoder model~\cite{transformer} in which the encoder-decoder model is pre-trained on large unlabeled data.

\subsection{Transformer-based Encoder-Decoder}

In this work, we define the Transformer-based encoder-decoder model 
as follows. 

\paragraph{Encoder}
The encoder consists of $M$ layer 
encoder blocks.
The input of the encoder is $X = \{x_i, x_2, ... x_L \}$. The output through the $M$ layer 
encoder blocks is defined as 
\begin{align}
H_e^M = \{h_{e1}^M, h_{e2}^M, ... h_{eL}^M \} \in \mathbb{R}^{L \times d}.
\end{align}
The 
encoder block consists of a self-attention module and a two-layer feed-forward network. 

\paragraph{Decoder}
The decoder consists of $M$ layer 
decoder blocks.
The inputs of the decoder are the output of the encoder $H_e^M$ and the output of the previous step of the decoder $\{ y_1,...,y_{t-1} \}$. The output through the $M$ layer Transformer decoder blocks is defined as 
\begin{align}
H_d^M = \{ h_{d1}^M,..., h_{dt}^M \} \in \mathbb{R}^{t \times d}.
\end{align}
 In each step $t$, the $h_{dt}^M$ is projected to 
 \textcolor{black}{the vocabulary space}
 and the decoder outputs the highest probability token as the next token. The Transformer decoder block consists of a self-attention module, a context-attention module, and a two-layer feed-forward network. 

\paragraph{Multi-head Attention}
The encoder and decoder
blocks use
multi-head attention,
which consists of a combination of $K$ attention
heads and is denoted as $\mathrm{Multihead}(Q, K, V) = \mathrm{Concat}(\mathrm{head}_1, ...,\mathrm{head}_k)W^o$, where each head is $\mathrm{head}_i = \mathrm{Attention}(QW_i^Q, KW_i^K , VW_i^V)$.

The weight matrix $A$ in each attention-head $\mathrm{Attention}(\tilde{Q}, \tilde{K}, \tilde{V}) = A \tilde{V}$ is defined as 
\begin{align}
A = \mathrm{softmax}\left(\frac{\tilde{Q}\tilde{K}^\top}{\sqrt{d_k}}\right) \in \mathbb{R}^{I \times J},
\label{eq:attn}
\end{align}
 where $d_k = d / k$, $\tilde{Q} \in \mathbb{R}^{I \times d}$, $\tilde{K}, \tilde{V} \in \mathbb{R}^{J \times d}$.

In the $m$-th layer of self-attention, the same representation $H^m_{\cdot}$ is given to $Q$, $K$,  and $V$. In the context-attention, we give $H^m_d$ to $Q$ and $H^M_e$ to $K$ and $V$. 

\subsection{Summary Loss Function}

To fine-tune the seq-to-seq model for abstractive summarization, we use cross entropy loss as 
\begin{align}
L_\mathrm{sum} = - \frac{1}{NT}\sum_{n=1}^N \sum_{t=1}^T \log P(y^n_{t}),
\end{align}
where $N$ is the number of training samples.

\section{Saliency Models}
Several studies have proposed the combination of a token-level saliency model and 
a 
seq-to-seq model,
\textcolor{black}{which is not pre-trained},  
and reported its effectiveness~\cite{bottomup, selective-encoding}. 
We 
also use a simple token-level saliency model 
\textcolor{black}{as a basic model }
in this study. 

\subsection{Basic Saliency Model}
A basic saliency model consists of $M$-layer Transformer encoder blocks ($\mathrm{Encoder}_\mathrm{sal}$) 
and a single-layer feed-forward network. 
We define the saliency score of the $l$-th token ($1 \leq l \leq L$) in the source text as 
\begin{equation}
\label{eq:sigmoid}
 S_l= \sigma(W_1^\top {\rm Encoder_{sal}}(X)_l+b_1), 
\end{equation}
 where ${\rm Encoder_{sal}()}$ represents the output of the last layer of 
 \textcolor{black}{$\rm Encoder_{sal}$},
 $W_1 \in \mathbb{R}^{d}$ and $b_1$ are learnable parameters, and $\sigma$ represents a sigmoid function. 

\subsection{Two Types of Saliency Model for Combination}
In this study, we use two types of saliency model for combination: a shared encoder and an extractor. Each model structure is based on the basic saliency model. We describe them below.

\paragraph{Shared encoder}
The shared encoder \textcolor{black}{shares the parameters of} \textcolor{black}{$\rm Encoder_{sal}$} and the encoder of the seq-to-seq model. This model is jointly trained with the seq-to-seq model and the saliency score is used to bias the representations of 
the seq-to-seq model. 
\paragraph{Extractor} The extractor extracts the important tokens or sentences from the source text on the basis of the saliency score. The extractor is separated with the seq-to-seq model, and each model is trained independently. 


\subsection{Pseudo Reference Label} 
\label{sec:pseudo_reference_label}

The saliency model predicts 
\textcolor{black}{the saliency score } 
$S_l$ for each token $x_l$. 
If there is a reference label $r_l$ $\in \{0, 1\}$ for each $x_l$, we can train the saliency model in a supervised manner. However, the reference label for each token is typically not given, since the training data for the summarization 
consists of only the source text and its reference summary. Although there are no reference saliency labels, we can make pseudo reference labels by aligning both source and summary token sequences and extracting common tokens~\cite{bottomup}. 
\textcolor{black}{We used pseudo labels when we train the saliency model in a supervised manner. }

\subsection{Saliency Loss Function}
\label{sec:saliency_loss}

To train the saliency model in a supervised way \textcolor{black}{with pseudo reference labels}, 
we use binary cross entropy loss 
 as 
\begin{gather}
L_\mathrm{sal} = - \frac{1}{NL} \sum_{n=1}^N \sum_{l=1}^L
\biggl\{
\begin{split}
&r^n_l \log S_l^n + \\
&(1-r^n_l) \log (1-S_l^n)
\end{split}
\biggr\},
\end{gather}
where $r_l^n$ is a pseudo reference label of token $x_l$ in the $n$-th sample.

\begin{figure*}[t]
    \centering
    \includegraphics[width=1.0\textwidth]{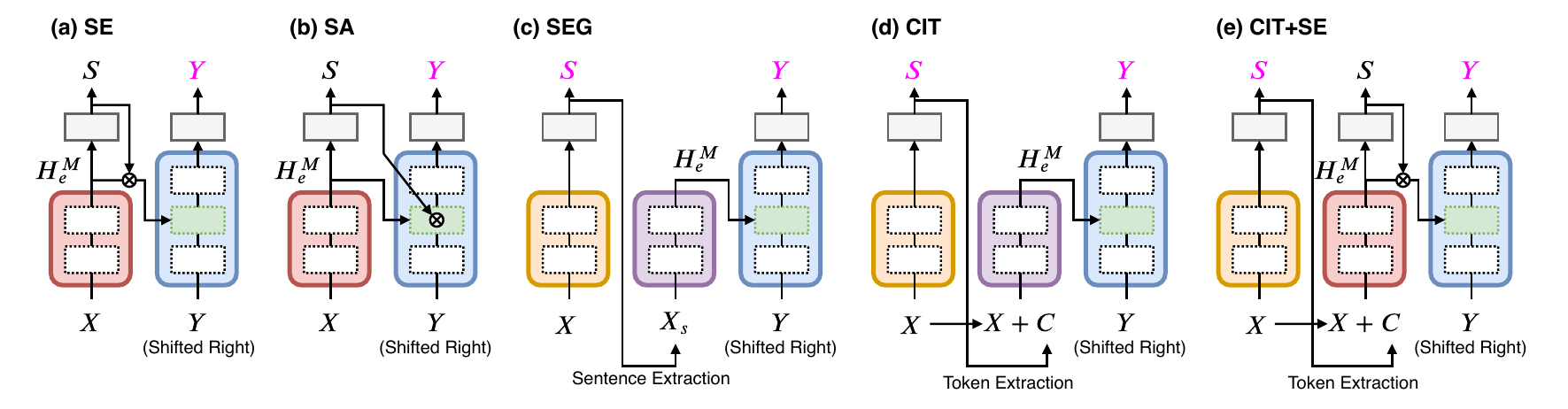}
     \caption{Combinations of seq-to-seq and saliency models.  
     \textbf{Purple}: Encoder.
     \textbf{Blue}: Decoder.
     \textbf{Red}: Shared encoder, which is a shared model for saliency detection and encoding, used in (a), (b), and (e).
     \textbf{Yellow}: Extractor, which is an independent saliency model to extract important (c) sentences $X_s$ or (d), (e) tokens $C$ from the source text $X$. 
     Each of these colored blocks represents $M$-layer Transformer blocks.
     \textbf{Gray}: Linear transformation.  
     \textbf{Green}: Context attention.
     \textbf{Pink}: Output trained in a supervised manner, where $S$ is the saliency score and $Y$ is the summary.}
     
    \label{fig:models}
\end{figure*}

\section{Combined Models}
This section describes
nine combinations of 
the pre-trained seq-to-seq model and saliency models. 

\paragraph{Combination types}
We roughly categorize the combinations into three types. 
Figure~\ref{fig:models} shows an image of each combination. 

The first type uses the shared encoder (\S\ref{sec:shared_model}). These models consist of the shared encoder and the decoder, where the shared encoder module 
\textcolor{black}{plays two roles: saliency detection and the encoding of the seq-to-seq model}. 
\textcolor{black}{The saliency scores are used to bias the representation of the seq-to-seq model for several models in this type.}

The second type uses the extractor (\S\ref{sec:seperated_model}, \S\ref{sec:seperated_model_cit}). These models consist of the extractor, encoder, and decoder and follow two steps: first, 
\textcolor{black}{the extractor}
\textcolor{black}{extracts}
the important tokens or sentences from the source text, and second, 
\textcolor{black}{the encoder uses them}
as an input of the seq-to-seq models. Our proposed model (CIT) belongs to this type. 

The third type uses both the shared encoder and the extractor (\S\ref{sec:shared_seperated_model}). These models consist of the extractor, shared encoder, and decoder and also follow two steps: first, 
\textcolor{black}{the extractor extracts}
the important tokens from the source text, and second, 
\textcolor{black}{the shared encoder uses}
them as an input of the seq-to-seq model.

\paragraph{Loss function}
From the viewpoint of the loss function, there are two major types of model: those that use the saliency loss (\S\ref{sec:saliency_loss}) and those that do not. 
We also denote the loss function for the seq-to-seq model as $L_\mathrm{abs}$ and the loss function for the extractor as $L_\mathrm{ext}$. 
\textcolor{black}{$L_\mathrm{ext}$ is trained with $L_\mathrm{sal}$, and $L_\mathrm{abs}$ is trained with $L_\mathrm{sum}$ or $L_\mathrm{sum} + L_\mathrm{sal}$. }


\subsection{Using Shared Encoder to Combine the Saliency Model and the  Seq-to-seq Model}
\label{sec:shared_model}
%


\paragraph{Multi-Task (MT)} 
This model trains the shared encoder and the decoder by minimizing both the summary and saliency losses. 
The loss function of this model is $L_\mathrm{abs} = L_\mathrm{sum} + L_\mathrm{sal}$.

\paragraph{Selective Encoding (SE)}
This model uses the saliency score to weight the shared encoder output. Specifically, the final output $h_{el}^M$ of the shared encoder is weighted as 
\begin{eqnarray}
\label{eq:selective_encoding}
 \tilde{h}_{el}^{M} = h_{el}^M S_l.
\end{eqnarray}
 Then, we replace the input of the decoder $h_{el}^M$ with $\tilde{h}_{el}^{M}$. Although \citet{selective-encoding} used BiGRU, we use Transformer for fair comparison. 
The loss function of this model is $L_\mathrm{abs} = L_\mathrm{sum}.$

\paragraph{Combination of SE and MT}
This model has the same structure as the SE.
The loss function of this model is $L_\mathrm{abs} = L_\mathrm{sum} + L_\mathrm{sal}$.

\paragraph{Selective Attention (SA)}
This model weights the attention scores of the decoder side, unlike the SE model. Specifically, the attention score $a_{i}^{t} \in \mathbb{R}^L$ in each step $t$ is weighted by $S_l$. $a_{i}^{t}$ is a $t$-th row of $A_i \in \mathbb{R}^{T \times L}$, which is a weight matrix of the $i$-th attention head in the context-attention (Eq.~\eqref{eq:attn}). 
\begin{eqnarray}
\label{eq:selective_attention}
   \tilde{a}_{il}^t = \frac{a_{il}^t  S_l}{\sum_l a_{il}^t S_l}.
\end{eqnarray}

\citet{bottomup} took a similar approach in that their model weights the copy probability of a pointer-generator model. However, as the pre-trained seq-to-seq model does not have a copy mechanism, we weight the context-attention for all Transformer decoder blocks. 
The loss function of this model is $L_\mathrm{abs} = L_\mathrm{sum}$.

\paragraph{Combination of SA and MT}
This model has the same structure as the SA.
The loss function of this model is $L_\mathrm{abs} = L_\mathrm{sum} + L_\mathrm{sal}$.


\subsection{Using the Extractor to Refine the Input Text}
\label{sec:seperated_model}

\paragraph{Sentence Extraction then Generation (SEG)}
This model first extracts the saliency sentences on the basis of a sentence-level saliency score $S_j$. $S_j$ is calculated by using the token level saliency score of the extractor,
$S_l$, as 
\begin{align}
S_j = \frac{1}{N_j} \sum_{l:x_l \in X_j} S_l,
\end{align}
where $N_j$ and $X_j$ are the number of tokens and the set of tokens within the $j$-th sentence. Top $P$ sentences are extracted according to the sentence-level saliency score and then concatenated as one text $X_s$. These extracted sentences are then used as the input of the seq-to-seq model.
 
In the training, we extracted $X_s$, which maximizes the ROUGE-L scores with the reference summary text. In the test, we used 
the average number of sentences in $X_{s}$ in the training set as~$P$. 
The loss function of the extractor is $L_\mathrm{ext} = L_\mathrm{sal}$, and that of the seq-to-seq model is $L_\mathrm{abs} = L_\mathrm{sum}$. 


\subsection{Proposed: Using Extractor to Extract an Additional Input Text}
\label{sec:seperated_model_cit}
\paragraph{Conditional Summarization Model with Important Tokens}
We propose a new combination of  
the extractor 
and the seq-to-seq model, CIT, which can consider important tokens explicitly. 
Although the SE and SA models softly weight the representations of the source text or attention scores, they cannot select salient tokens explicitly. 
SEG explicitly extracts the salient sentences from the source text, but it cannot give token-level information to the seq-to-seq model, and it sometimes drops important information when extracting 
sentences. 
In contrast, CIT 
uses the tokens extracted according to saliency scores as an \textit{additional} input of the seq-to-seq model. 
By adding token-level information, CIT can effectively guide the abstractive summary without dropping any important information.

Specifically, $K$ tokens $C = \{c_1, ..., c_K \}$ are extracted in  descending order of the saliency score $S$. $S$ is obtained by inputting $X$ to the extractor.  
The order of $C$ retains the order of the source text $X$. A combined text $\tilde{X} = \mathrm{Concat}(C, X)$ is given to the seq-to-seq model as the input text. 
The loss function of the extractor is $L_\mathrm{ext} = L_\mathrm{sal}$, and that of the seq-to-seq  model is $L_\mathrm{abs} = L_\mathrm{sum}$. 


\subsection{Proposed: Combination of Extractor and Shared Encoder}
\label{sec:shared_seperated_model}

\paragraph{Combination of CIT and SE}
This model combines the CIT and SE, so 
CIT uses an extractor
for extracting important tokens, and 
SE is trained by using a shared encoder in the seq-to-seq model.
The SE model is trained in an unsupervised way.
The output $H^M_e \in \mathbb{R}^{L+K}$ of 
the shared encoder
is weighted by saliency score $S \in \mathbb{R}^{L+K}$ with Eq.~\eqref{eq:selective_encoding}, where $S$ is estimated by using the output of 
the shared encoder with Eq.~\eqref{eq:sigmoid}.
The loss function of the extractor is $L_\mathrm{ext} = L_\mathrm{sal}$, and that of the seq-to-seq model is $L_\mathrm{abs} = L_\mathrm{sum}$. 

\paragraph{Combination of CIT and SA}
This model combines the CIT and SA, so we also train two saliency models. 
The SA model is trained in an unsupervised way, the same as the CIT + SE model.  The attention score $a_i^t \in \mathbb{R}^{L+K}$ is weighted by $S \in \mathbb{R}^{L+K}$ with Eq.~\eqref{eq:selective_attention}.
The loss function of the extractor is $L_\mathrm{ext} = L_\mathrm{sal}$, and that of the seq-to-seq model is $L_\mathrm{abs} = L_\mathrm{sum}$. 

\section{Experiments}
\subsection{Dataset}
We used the CNN/DM dataset~\cite{cnndm} and the XSum dataset~\cite{xsum}, which are both standard datasets for news summarization. The details of the two datasets are listed in Table~\ref{tb:dataset}. The CNN/DM is a highly extractive summarization dataset and the XSum is a highly 
abstractive 
summarization dataset.

\subsubsection{Model Configurations}
\textcolor{black}{We used BART$_{\mathrm {LARGE}}$~\cite{bart}, which is one of the state-of-the-art models, as the pre-trained seq-to-seq model and
RoBERTa$_\mathrm{BASE}$~\cite{roberta} as the initial model of the extractor.  In the extractor of CIT, stop words and duplicate tokens are ignored for the XSum dataset.
}

We used 
fairseq\footnote{\label{note:fairseq}\url{https://github.com/pytorch/fairseq}} for the implementation of the seq-to-seq model.
For fine-tuning of BART$_\mathrm{LARGE}$ and the combination models, we used the same parameters as the official code\label{note:fairseq}. 
For fine-tuning of RoBERTa$_\mathrm{BASE}$, 
we used Transformers\footnote{\label{note:transformers}\url{https://github.com/huggingface/transformers}}. We set the learning rate to 0.00005 and the batch size to 32.  

\subsection{Evaluation Metrics}
We used ROUGE scores (F1), including ROUGE-1 (R-1), ROUGE-2 (R-2), and ROUGE-L (R-L), as the evaluation metrics~\cite{Lin:2004}. ROUGE scores were calculated using the files2rouge toolkit\footnote{\url{https://github.com/pltrdy/files2rouge}}.
 
\begin{table}[t]
\centering
{\small \tabcolsep=3pt
\begin{tabular}{c|cccc}
\hline
set   & train & dev & eval & avg. length\\ \hline
CNN/DM& 287,227 & 13,368 & 11,490 & 69.6 \\
XSum & 203,150 & 11,279 & 11,267 & 26.6\\
\hline
\end{tabular}
}
\caption{Details of the datasets used in this paper.\label{tb:dataset}}
\end{table}

\subsection{Results}
\paragraph{Do saliency models improve summarization accuracy in highly \textit{extractive} datasets?}
Rouge scores of the combined models on the CNN/DM dataset are shown in Table~\ref{tab:result_main_cnndm}. 
We can see that all combined models outperformed the simple fine-tuned BART. This indicates that the saliency detection is effective in highly extractive datasets.
One of the proposed models, CIT + SE, achieved the highest accuracy. 
The CIT model alone also 
outperformed other saliency models.
This indicates that the CIT model effectively guides the abstractive summarization by combining explicitly extracted tokens. 

\begin{table}[t]
    \centering
    \small 
    \begin{tabular}{l|ccc} \hline
         models & R1 & R2 & RL \\ \hline
         BART~\cite{bart} & 44.16 & 21.28 & 40.90 \\ 
         BART (our fine-tuning) & 43.79 & 21.00 & 40.58 \\ \hline
         MT & 44.84  & 21.71 & 41.52 \\
         SE  & 44.59 & 21.49 & 41.28 \\
         SE + MT & 45.23 & 22.07 & 41.94 \\
         SA  & 44.72 & 21.59 & 41.40 \\
         SA + MT & 44.93 & 21.81 & 41.61 \\ \hline
         SEG & 44.62 & 21.51 & 41.29 \\ \hline
         CIT &  45.74 & 22.50 & 42.44 \\ \hline
         CIT + SE & \bf 45.80 & \bf 22.53 & \bf 42.48 \\
         CIT + SA &  45.74 & 22.48 & 42.44 \\ 
         \hline
    \end{tabular}
    \caption{Results of BART and combined models on CNN/DM dataset. 
    Five row-groups are 
    the models described in \S\ref{sec:seq-to-seq}, 
    \S\ref{sec:shared_model}, \S\ref{sec:seperated_model},
    \S\ref{sec:seperated_model_cit}, and \S\ref{sec:shared_seperated_model} in order from top to bottom.}
    
　　\label{tab:result_main_cnndm}
\end{table}
\begin{table}[t]
    \centering
    \small 
    \begin{tabular}{l|ccc} \hline
         models & R-1 & R-2 & R-L \\ \hline
         BART~\cite{bart} & 45.14 & \bf 22.27 & \bf 37.25 \\ 
         BART (our fine-tuning) & 45.10 & 21.80 & 36.54 \\ \hline
         MT & 44.57 & 21.40 & 36.31 \\
         SE & 45.34 & 21.98 & 36.79 \\
         SE + MT & 44.64 & 21.30 & 36.07 \\
         SA & 45.35 & 22.02 & 36.83 \\
         SA + MT & 45.37 & 21.98 & 36.77 \\ \hline
         SEG & 41.03 & 18.20 & 32.75 \\ \hline
         CIT  & \bf 45.42 & \underline{22.13} & \underline{36.92} \\ \hline
         CIT + SE & 45.36 & 22.02 & 36.83 \\
         CIT + SA & 45.30 & 21.94 & 36.76 \\
         \hline
    \end{tabular}
    \caption{Results of BART and combined models 
    on XSum dataset.
    The underlined result represents the best result among the models that outperformed our simple fine-tuning result.}
    
    \label{tab:result_main_xsum}
\end{table}

\paragraph{Do saliency models improve summarization accuracy in highly \textit{abstractive} datasets?}

Rouge scores of the combined models on the XSum dataset are shown in Table~\ref{tab:result_main_xsum}. 
The CIT model performed the best,
although its improvement was smaller than on the CNN/DM dataset.
Moreover, the accuracy of the MT, SE + MT, and SEG models 
decreased on the XSum dataset. 
These results were very different from those on the CNN/DM dataset. 

One reason for the difference can be traced to  
the quality of the pseudo saliency labels. 
CNN/DM is a highly extractive dataset, so it is relatively easy to create token alignments for generating pseudo saliency labels, while in contrast, a summary in XSum is highly abstractive and short, which makes it difficult to create pseudo labels with high quality by simple token alignment. 
To improve the accuracy of summarization in this dataset, we have to improve the quality of the pseudo saliency labels and the
accuracy of the saliency model.


\paragraph{How accurate are the outputs of the extractors?}

\begin{table}[t]
    \centering
    \small 
    \tabcolsep=3pt
    \begin{tabular}{l|ccc|ccc} \hline
         & \multicolumn{3}{c}{CNN/DM} & 
         \multicolumn{3}{|c}{XSum}  \\ \hline
         models & R-1 & R-2 & R-L 
         & R-1 & R-2 & R-L  \\ \hline
         Lead3 & 40.3 & 17.7 & 36.6 & 16.30 & 1.61 & 11.9 \\ 
         BertSumExt$^1$ & 43.25 & 20.24 & \bf 39.63 & -- & -- & -- \\ 
         CIT: Top-$K$ toks & \bf 46.72 & \bf 20.53 & 37.73 & \bf 24.28 & 3.50 & 15.63 \\ 
         CIT: Top-3 sents &  42.64 &  20.05 & 38.89  & 21.71 & \bf 4.45 & \bf 16.91 \\ \hline
    \end{tabular}
    \caption{Results of saliency models on CNN/DM and XSum datasets. CIT extracted the top-$K$ tokens or top-$3$ sentences from the source text. $^1$\cite{presum}.
    The summaries in XSum are highly extractive, so the result of BertSumExt for XSum was not reported.}
    
    \label{tab:result_saliency}
\end{table}

We analyzed the quality of the tokens extracted by the extractor in CIT. 
The results are summarized in Table~\ref{tab:result_saliency}. 
On the CNN/DM dataset, the ROUGE-1 and ROUGE-2 scores of our extractor (Top-$K$ tokens) were higher than other models,  
while the ROUGE-L score was lower than the other sentence-based extraction method.
This is because that our token-level extractor finds the important tokens whereas the seq-to-seq model learns how to generate a fluent summary incorporating these important tokens. 

On the other hand, the extractive result on the XSum dataset was lower. For highly abstractive datasets, there is little overlap between the tokens. 
We need to consider how to make the high-quality pseudo saliency labels and how to evaluate the similarity of these two sequences.


\paragraph{Does the CIT model outperform other fine-tuned models?}

\begin{table}[t]
    \centering
    \small 
    \tabcolsep=3pt
    \begin{tabular}{l|cc|ccc} \hline
         models & data & params &  R-1 & R-2 & R-L \\ \hline
         MASS$_\mathrm{BASE}$$^1$ & 16G & 123M & 42.12 & 19.50 & 39.01 \\
         BertSumExtAbs$^2$ & 16G & 156M & 42.13 & 19.60 & 39.18 \\
         UniLM$^3$ & 16G & 340M & 43.33 & 20.21 & 40.51 \\ 
         T5$_\mathrm{11B}$$^4$ & 750G & 11B &  43.52 & 21.55 & 40.69 \\ 
         BART$_\mathrm{LARGE}^5$ & 160G & 400M & 44.16 & 21.28 & 40.90\\ 
         PEGASUS$_\mathrm{C4}^6$ & 750G & 568M & 43.90 & 21.20 & 40.76 \\ 
         PEGASUS$_\mathrm{HugeNews}^6$ & 3.8T & 568M & 44.17 & 21.47 & 41.11 \\ 
         ProphetNet$^7$ & 160G & 400M & 44.20 & 21.17 & 41.30 \\ 
         ERNIE-GEN$_\mathrm{LARGE}^8$ & 16G & 340M & 44.02 & 21.17 & 41.26 \\ 
         UniLMv2$_\mathrm{BASE}$$^9$ & 160G & 110M & 43.16 & 20.42 & 40.14 \\ 
         \hline
         CIT  & 160G & \textcolor{black}{525}M  & \bf 45.74 & \bf 22.50 & \bf 42.44 \\
         \hline
    \end{tabular}
    \caption{Results of state-of-the-art models and the proposed model on CNN/DM dataset. 
    We also report the size of pre-training data and parameters
    utilized for each model. 
    $^1$\cite{mass}; $^2$\cite{presum}; $^3$\cite{unilm};
    $^4$\cite{2019t5}; $^5$\cite{bart}; $^6$\cite{pegasus}
    $^7$\cite{prophetnet}; $^8$\cite{ernie-gen}; $^9$\cite{unilmv2}}
    \label{tab:result_main2_cnndm}
\end{table}

\begin{table}[t]
    \centering
    \small 
    \tabcolsep=3pt
    \begin{tabular}{l|cc|ccc} \hline
         models & data & params & R-1 & R-2 & R-L \\ \hline
         MASS$_\mathrm{BASE}$$^1$ & 16G & 123M
         & 39.75 & 17.24 & 31.95 \\ 
         BertSumExtAbs$^2$ & 16G & 156M &  38.81 & 16.50 & 31.27 \\
         BART$_\mathrm{LARGE}^3$ & 160G & 400M & 45.14 & 22.27 & 37.25 \\ 
         PEGASUS$_\mathrm{C4}^4$ & 750G & 568M & 45.20 & 22.06 & 36.99 \\
         PEGASUS$_\mathrm{HugeNews}^4$ & 3.8T & 568M & \bf 47.21 & \bf 24.56 & \bf 39.25 \\ 
         UniLMv2$_\mathrm{BASE}$$^5$ & 160G & 110M & 44.00 & 21.11 & 36.08 \\
         \hline
         CIT & 160G & \textcolor{black}{525}M  & 45.42 &  22.13 & 36.92 \\
         \hline
    \end{tabular}
    \caption{Results of state-of-the-art models and the proposed model on XSum dataset.
    $^1$\cite{mass}; $^2$\cite{presum}; $^3$\cite{bart}; $^4$\cite{pegasus}; $^5$\cite{unilmv2} 
    }
    \label{tab:result_main2_xsum}
\end{table}

Our study focuses on the combinations of saliency models and the pre-trained seq-to-seq model. However, there are several studies that focus more on the pre-training strategy. We compared the CIT model with those models. 
Their ROUGE scores are shown in Tables~\ref{tab:result_main2_cnndm} 
and~\ref{tab:result_main2_xsum}. 
From Table~\ref{tab:result_main2_cnndm}, we can see that our model outperformed the recent pre-trained models on the CNN/DM dataset. Even though PEGASUS$_\mathrm{HugeNews}$ was pre-trained on 
the largest corpus comprised of news-like articles, the accuracy of abstractive summarization 
was not improved
much. 
Our model improved the accuracy without any additional pre-training. 
This result indicates that it is more effective to combine saliency models with the 
seq-to-seq model for generating a highly extractive summary. 

On the other hand, on the XSum dataset, PEGASUS$_\mathrm{HugeNews}$ improved the ROUGE scores and achieved the best results. In the XSum dataset, summaries often include the expressions that are not written in the source text. Therefore, increasing the pre-training data and learning more patterns were effective. However, by improving the quality of the pseudo saliency labels, we should be able to improve the accuracy of the CIT model.   


\section{Related Work and Discussion}
\paragraph{Pre-trained Language Models for Abstractive Summarization}
\citet{Bertsum} used BERT for their sentence-level extractive summarization model. \citet{HIBERT} proposed a new pre-trained model that considers document-level information for sentence-level extractive summarization. Several researchers have published pre-trained encoder-decoder models very recently~\cite{poda,bart,2019t5}. \citet{poda} pre-trained a Transformer-based pointer-generator model. \citet{bart} pre-trained a standard Transformer-based encoder-decoder model using large unlabeled data and achieved state-of-the-art results. \citet{unilm} and \citet{ernie-gen} extended the BERT structure to handle seq-to-seq tasks. 

All the studies above focused on how to learn a universal pre-trained model; they did not consider the combination of pre-trained and saliency models for an abstractive summarization model.  

\paragraph{Abstractive Summarization with Saliency Models}
\citet{unified}, \citet{bottomup}, and \citet{ETADS} incorporated a sentence- and word-level extractive model in the pointer-generator model. Their models weight the copy probability for the source text by using an extractive model and guide the pointer-generator model to copy important words. \citet{keyword_guided} proposed a keyword guided abstractive summarization model. \citet{sentence_rewriting} proposed a sentence extraction and re-writing model that is trained in an end-to-end manner by using reinforcement learning. \citet{retrieve_rewrite} proposed a search and rewrite model. \citet{EXCONSUMM_Compressive} proposed a combination of sentence-level extraction and compression. None of these models are based on a pre-trained model. In contrast,  our purpose is to clarify whether combined models are effective or not, and we are the first to investigate the combination of pre-trained seq-to-seq and saliency models. We compared a variety of combinations and clarified which combination is the most effective. 
 
\section{Conclusion}
This is the first study that has conducted extensive experiments to investigate the effectiveness of incorporating saliency models into the pre-trained seq-to-seq model. 
From the results, we found that saliency models were effective in finding important parts of the source text, even if the seq-to-seq model is pre-trained on large-scale corpora, especially for generating an highly extractive summary.
We also proposed a new combination model, CIT, that outperformed simple fine-tuning and other combination models. 
Our combination model improved the summarization accuracy without any additional pre-training data and can be applied to any pre-trained model. 
While recent studies have been conducted to improve summarization accuracy by increasing the amount of pre-training data and developing new pre-training strategies, this study sheds light on the importance of saliency models in abstractive summarization.

\bibliography{emnlp-ijcnlp-2019}
\bibliographystyle{acl_natbib}

\end{document}